\RequirePackage[OT1]{fontenc}
\documentclass[letterpaper,10pt,journal,twoside]{IEEEtran}
\IEEEoverridecommandlockouts
\usepackage{amssymb, mathtools, xargs, tensor, units, cite, mathrsfs, algpseudocode, graphicx, xcolor, amsmath, listings, algorithm, tikz, pgfplots, cancel, multicol, geometry, multirow, booktabs,bm,booktabs,hyperref}
\usepackage{placeins}
\usepackage[table]{xcolor}
\usepackage{booktabs}
\usepgfplotslibrary{fillbetween} 
\algnewcommand\algorithmicforeach{\textbf{for each}}
\algdef{S}[FOR]{ForEach}[1]{\algorithmicforeach\ #1\ \algorithmicdo}
\usepackage{pgfplots}
\usepackage{pgfplotstable}
\usepackage{amsmath}
\pgfplotsset{compat=newest}
\usepgfplotslibrary{groupplots}
\pgfplotsset{compat=1.17}
\usepgfplotslibrary{colormaps}
\definecolor{customgreen}{rgb}{0.0, 0.6, 0.0}
\usepackage{soul}

\pgfplotsset{compat=1.18}
\usetikzlibrary{3d, calc}
\pgfplotsset{compat=newest}
\usetikzlibrary{arrows.meta}

\newtheorem{assumption}{Assumption}

\newtheorem{remark}{Remark}

\newcommand{\B}[1]{{\color{black}#1}}

\allowdisplaybreaks
\title{\LARGE Input Bias Correction in Gaussian Process Regression}
\author{Muzaffar Qureshi$^{1}$ \and Tochukwu E. Ogri$^{1}$  \and Trivikram Satharasi$^{1}$ \and Kyle Volle$^{2}$ \and Rushikesh Kamalapurkar$^{1}$
\thanks{This research was supported in part by the Air Force Research Laboratory under contract number FA8651-24-1-0019. Any opinions, findings, or recommendations in this article are those of the author(s), and do not necessarily reflect the views of the sponsoring agencies.}%
\thanks{$^{1}$ Department of Mechanical and Aerospace Engineering, University of Florida, Gainesville, FL, USA, email: {\tt\footnotesize \{muzaffar.qureshi, tochukwu.ogri, t.satharasi, rkamalapurkar\} @ufl.edu}.}%
\thanks{$^{2}$ Torch Technologies, Shalimar, FL, USA, email: {
\tt \footnotesize kyle.volle@torchtechnologies.com}}
\thanks{%
This article was accepted for publication in
\emph{IEEE Robotics and Automation Letters} (Oct 2026).
\textcopyright~2026 IEEE.
Personal use of this material is permitted.
}%

}

\begin{document}
\maketitle
\pagestyle{empty}

\begin{abstract}
In traditional Gaussian process (GP) regression, the inputs are typically assumed to be known, and uncertainty enters only through the measurements. When scalar fields are mapped using GP regression with measurements obtained from a mobile sensor, the inputs correspond to the estimated pose of the mobile sensor. The goal in this letter is to understand how a recorded GP, trained on erroneous inputs (e.g., pose estimates from dead-reckoning), can be corrected in a numerically efficient way when better estimates of the inputs become available. In the developed bias-aware GP (Ba-GP) framework, the recorded GP posterior-mean gradient is used to propagate estimated input bias and noise into mean and covariance corrections. The resulting additive covariance structure is then exploited to develop numerically efficient first-order correction algorithms. A simulation study and an indoor light-intensity mapping experiment show that Ba-GP improves predictive accuracy and uncertainty calibration relative to noisy-input GP baselines with an order-of-magnitude reduction in computational cost relative to full GP retraining.
\end{abstract}

\section{Introduction}\label{sec:introduction}
Gaussian processes (GPs) provide a Bayesian framework for learning scalar fields from sparse measurements \cite{SCC.Rasmussen.Williams2006, qureshi2025gaussianprocessbasedscalarfield}, yielding predictive means and uncertainty estimates. \B{Conventional GP regression assumes that the input locations are known accurately; however, this assumption may not hold in real-world applications. The motivating setting of this letter is mobile sensing in which measurements are incorporated into a GP model as they are collected, but accurate measurement locations are not available at a high frequency. High-frequency odometry data from sensors such as IMUs and wheel encoders provide location estimates in real-time but accumulate drift \cite{SCC.Liu.Mao.ea2025,Qureshi2025FaultTolerant}, whereas lower-bias absolute-position measurements, such as GPS or motion-capture measurements, are available only intermittently. Consequently, waiting for the final refined trajectory before training would preclude the use of the GP map during the sensing episode. The GP is therefore trained using the available odometry-based locations whose errors can degrade the resulting map \cite{SCC.GhaffariJadidi.Miro.ea2017,SCC.Lilienthal.Duckett2004, TCST}. After the episode, filtering or smoothing techniques can be used to refine the recorded trajectory and reduce the accumulated localization drift \cite{SCC.Qureshi.Ogri.ea2025}. The GP could then be retrained using the refined locations; however, doing so requires rebuilding the GP using the corrected input data.}

Existing uncertain-input GP methods primarily address stochastic uncertainty in the inputs. For multi-step-ahead forecasting, Girard \emph{et al.} \cite{SCC.Girard.Rasmussen.ea2002,SCC.Girard2004} approximated the non-Gaussian predictive distribution induced by uncertain regressors with a moment-matched Gaussian distribution. Dallaire \emph{et al.} \cite{SCC.Dallaire.Besse.ea2009} showed that, for independent Gaussian input locations and a squared-exponential kernel, the covariance matrix can be computed analytically, yielding a modified GP kernel. McHutchon and Rasmussen \cite{SCC.McHutchon.Rasmussen2011} developed a first-order Taylor approximation, termed noisy-input GP (NIGP), in which measurement-location uncertainty is represented as an additional location-dependent noise term proportional to the squared gradient of the posterior mean.  \B{These methods account for the noise in the input locations, but they do not use an estimated input bias to correct a GP that has already been trained on recorded inputs.}

\B{In a recent study \cite{SCC.Qureshi.Ogri.ECCea2026}, input bias is corrected by treating the GP posterior mean and covariance as functions of the training inputs and applying Taylor expansions to compute Jacobians and Hessians offline. The resulting derivatives can then be used to correct the GP posterior when the bias estimate becomes available. However, the method does not account for the remaining input noise. In addition, the memory required to store the Jacobian and Hessian tensors increases with the dataset size, and the derivatives must be recomputed when predictions are required at new test inputs.}

\B{A straightforward way to address input bias is to retrain the GP with bias-corrected input locations. However, corrected input retraining requires reconstructing and factorizing the GP covariance matrix, whose computational cost scales cubically with the dataset size. When a GP has already been trained using recorded inputs, retraining also does not reuse the matrix computations already available from that GP.}

\B{These limitations motivate a GP correction framework that leverages an estimated input bias to update an already trained GP, explicitly accounts for the remaining input noise, and avoids complete retraining. To address this challenge, a bias-aware Gaussian process (Ba-GP) framework is developed. Ba-GP first uses the recorded-input GP to compute local gradients and propagates the estimated input bias through these gradients to obtain a first-order correction to the predictive mean. The residual input noise is then incorporated through an additive covariance correction. Conditioning on the corrected mean and covariance yields the Ba-GP posterior distribution. Furthermore, the additive covariance structure enables first- and second-order Neumann approximations that reuse the recorded GP matrix computations, thereby avoiding direct factorization of the full Ba-GP covariance matrix and reducing computational cost.} The contributions of this letter are as follows:
\begin{enumerate}
    \B{\item A bias-aware GP framework, Ba-GP, is developed for regression with input bias and input noise. The Ba-GP posterior separates bias from noise through a first-order mean correction and an additive covariance correction to the recorded-input covariance matrix.
    \item First- and second-order Neumann approximations of the Ba-GP posterior are derived by exploiting the additive covariance correction, reusing the recorded GP factorization to reduce computational cost while retaining accuracy.
    \item The Ba-GP framework and its Neumann approximations are evaluated in simulation and mobile robot mapping experiments, demonstrating improved accuracy under localization drift and characterizing the accuracy--computation tradeoff.}
\end{enumerate}

\section{Problem Formulation}
\label{sec:problem_formulation}
Consider a sensor-equipped mobile robot operating in a compact domain \(\mathcal{X}\subset\mathbb{R}^{d}\) that is tasked with estimation of an unknown scalar field \(h:\mathcal{X}\rightarrow\mathbb{R}\). Let \(N\) denote the total number of field measurements. The \(i\)-th measurement acquired at time \(t_i\) is associated with a true input \(\mathbf{x}(t_i):=\mathbf{x}_{i}\in\mathcal{X}\), and is given by
\begin{equation}
y_{i}=h(\mathbf{x}_{i})+\epsilon_{y,i},
\label{eq:pf_measurement_model}
\end{equation}
where \(\epsilon_{y,i}\sim\mathcal{N}(0,\sigma_{y}^{2})\) denotes independent measurement noise. 

Let \(X\coloneqq\{\mathbf{x}_{i}\}_{i=1}^{N}\) denote the set of true inputs. Assume that $X$ is unavailable during data collection (due to, for example, localization drift). The robot thus initially associates each measurement with a recorded input $\hat{\mathbf{x}}_i$ (obtained, for example, from dead reckoning). That is, the GP is initially trained using the available but recorded dataset
\begin{equation}
\mathcal{D}^{c}\coloneqq\{(\hat{\mathbf{x}}_i,y_i)\}_{i=1}^{N},
\label{eq:pf_corrupted_dataset}
\end{equation}
instead of the ideal but unavailable dataset
\begin{equation}
\mathcal{D}\coloneqq\{(\mathbf{x}_i,y_i)\}_{i=1}^{N}.
\label{eq:pf_true_dataset}
\end{equation}

Assume that after the recorded GP has been trained, intermittent high-fidelity measurements (e.g., from a GPS receiver) are processed via a state estimator to infer a posterior distribution over the true inputs. The posterior distribution is modeled as a Gaussian random variable $$\mathbf{x}_i \sim \mathcal{N}(\bar{\mathbf{x}}_i, \Sigma_{x,i}),$$
where the posterior mean $\bar{\mathbf{x}}_i = \hat{\mathbf{x}}_i + \hat{\delta}_i$ shifts the recorded input by an estimated bias correction $\hat{\delta}_i$, and the covariance matrix $\Sigma_{x,i} \in \mathbb{R}^{d \times d}$ encapsulates the total residual uncertainty, including both the estimator's uncertainty and the underlying stochastic sensor noise.

\begin{remark}\label{rem:independence}
    In practice, the total input covariance $\Sigma_{x,i} \in \mathbb{R}^{d \times d}$ must account for both the uncertainty of the bias estimator (e.g., the covariance output by an RTS smoother, denoted $\Sigma_{\delta,i}$) and the underlying high-frequency stochastic noise of the local sensors (denoted $\Sigma_{\epsilon,i}$). While the instantaneous sensor noise and the smoothed bias estimate generally exhibit cross-correlations, tracking the full cross-correlation history between the smoothed state and the instantaneous measurements is algorithm-dependent and generally computationally intractable \cite{SCC.Julier.Uhlmann1997}. Furthermore, due to the timescale separation between the slowly varying localization drift and the high-frequency sensor jitter, the assumption that the conditional posterior distribution of the bias is independent of the sensor noise is reasonable. Under this independence assumption, the total uncertainty is strictly additive: $\Sigma_{x,i} = \Sigma_{\delta,i} + \Sigma_{\epsilon,i}$.
\end{remark}

Equivalently, the posterior input can be represented as
\begin{equation}
\mathbf{x}_i
=
\hat{\mathbf{x}}_i+\hat{\delta}_i+\epsilon_{x,i},
\qquad
\epsilon_{x,i}\sim\mathcal{N}(\mathbf{0},\Sigma_{x,i}),
\label{eq:pf_posterior_input_model}
\end{equation}
where $\epsilon_{x,i}$ denotes the total residual input error about
the posterior mean $\bar{\mathbf{x}}_i$.
\B{The objective of this letter is to develop a GP correction framework that uses refined estimates of the inputs, when available, to correct the posterior distribution learned from the recorded dataset \(\mathcal{D}^{c}\) without the need for computationally expensive GP-retraining.}

\subsection{GP Regression}
\label{sec:gp_regression}
GP regression is a nonparametric Bayesian method for nonlinear function approximation \cite{SCC.Rasmussen.Williams2006}. The scalar field \(h\) is modeled as $h \sim \mathcal{GP}\big(0,\,k(\mathbf{x},\mathbf{x}')\big)$, where \(k:\mathcal{X}\times\mathcal{X}\rightarrow\mathbb{R}\) is the covariance kernel, and a zero prior mean is assumed. In this work, the squared-exponential kernel is adopted
\begin{equation}
k(\mathbf{x},\mathbf{x}')
=
\sigma_f^2
\exp\left(
-\frac{\|\mathbf{x}-\mathbf{x}'\|^2}{2\ell^2}
\right),
\label{eq:se_kernel_rw}
\end{equation}
where \(\sigma_f^2>0\) denotes the signal variance and \(\ell>0\) denotes the characteristic length-scale. Since the true input locations \(\mathbf{x}_i\) are unavailable online, GP regression uses the recorded dataset \(\mathcal{D}^{c}\) in \eqref{eq:pf_corrupted_dataset}. \B{The recorded input locations and corresponding field measurements are collected as}
\begin{equation*}
\scalebox{0.9}{$\B{\hat{\mathbf{X}}} \coloneqq
\begin{bmatrix}
\hat{\mathbf{x}}_{\B{1}}^{\top}\dots
\hat{\mathbf{x}}_{\B{N}}^{\top}
\end{bmatrix}^{\top}\in\mathbb{R}^{\B{N}\times d}, \quad
\B{\mathbf{Y}} \coloneqq
\begin{bmatrix}
y_{\B{1}}\dots
y_{\B{N}}
\end{bmatrix}^{\top}\in\mathbb{R}^{N},$}
\end{equation*}
and the corresponding recorded dataset is defined in \eqref{eq:pf_corrupted_dataset}. Let \(\B{K^{c}}\in\mathbb{R}^{\B{N \times N}}\) denote the covariance matrix formed from the recorded inputs, with entries
\begin{equation}
\B{[K^{c}]_{ij} = k(\hat{\mathbf{x}}_{i}, \hat{\mathbf{x}}_{j}), \qquad i,j=1,\dots,N,}
\label{eq:gp_corrupted_batch_kernel}
\end{equation}
and define $\B{K^{c,y}}\coloneqq \B{K^{c}}+\sigma_y^2 \B{I_{N}}$, where \(\B{I_{N}}\) denotes the \(\B{N \times N}\) identity matrix. For a query point \(\mathbf{x}_\ast\in\mathcal{X}\), define $\B{\mathbf{k}^{c}}(\mathbf{x}_\ast)
\coloneqq
\bigl[k(\hat{\mathbf{x}}_{i},\mathbf{x}_\ast)\bigr]_{i=1}^{\B{N}}\in\mathbb{R}^{\B{N}}$. The resulting joint Gaussian distribution is
\begin{equation}
\begin{bmatrix}
\B{\mathbf{Y}} \\
h(\mathbf{x}_\ast)
\end{bmatrix}
\sim
\mathcal{N}\left(
\begin{bmatrix}
\mathbf{0} \\
0
\end{bmatrix},
\begin{bmatrix}
\B{K^{c,y}} & \B{\mathbf{k}^{c}}(\mathbf{x}_\ast) \\
\B{\mathbf{k}^{c}}(\mathbf{x}_\ast)^{\top} & k(\mathbf{x}_\ast,\mathbf{x}_\ast)
\end{bmatrix}
\right),
\label{eq:gp_joint_corrupted}
\end{equation}
and conditioning on \(\B{\mathcal{D}^{c}}\) yields
\begin{equation}
\B{\mu^{c}}(\mathbf{x}_\ast)
=
\B{\mathbf{k}^{c}}(\mathbf{x}_\ast)^{\top}(\B{K^{c,y}})^{-1}\B{\mathbf{Y}},
\label{eq:gp_mean_corrupted}
\end{equation}
\begin{equation}
(\B{\sigma^{c}}(\mathbf{x}_\ast))^{2}
=
k(\mathbf{x}_\ast,\mathbf{x}_\ast)
-
\B{\mathbf{k}^{c}}(\mathbf{x}_\ast)^{\top}(\B{K^{c,y}})^{-1}\B{\mathbf{k}^{c}}(\mathbf{x}_\ast).
\label{eq:gp_variance_corrupted}
\end{equation}
If the ideal but unavailable dataset \(\mathcal{D}\) in \eqref{eq:pf_true_dataset} were used instead, the corresponding posterior mean and variance are denoted by \(\mu(\cdot)\) and \((\sigma(\cdot))^{2}\). \B{Retraining the final GP on the bias-corrected input locations would require inverting a $N\times N$ covariance matrix, incurring a $\mathcal{O}(N^3)$ computational cost \cite{SCC.Rasmussen.Williams2006}. Section~\ref{sec:retrospective_cost}
shows that the exact Ba-GP correction shares this leading order and develops approximations to reduce the cost to $\mathcal{O}(N^2)$.}

\section{Bias Correction in GP Models}
\label{sec:bias_gp_derivation}
This section shows how the estimated bias \(\hat{\delta}_i\) contributes a first-order correction to the predictive mean, while the residual noise term \(\epsilon_{x,i}\) introduces an additional variance component. From
\eqref{eq:pf_posterior_input_model}, the \(i\)-th field measurement is given as
\begin{equation}
y_i =
h\big(\hat{\mathbf{x}}_i+\B{\hat{\delta}_i}+\epsilon_{x,i}\big)
+
\epsilon_{y,i},
\qquad i=1,\dots,N.
\label{eq:bias_observation_compact}
\end{equation}
Using the zero-mean properties of \(\epsilon_{x,i}\) and \(\epsilon_{y,i}\), and treating these noise terms as independent of the GP prior, \B{we assume that the residual input and measurement noises are mutually uncorrelated and uncorrelated across measurements, i.e.,}
\begin{equation}\scalebox{0.95}{$
\mathrm{Cov}(\epsilon_{x,i},\epsilon_{x,j})=\mathbf{0}_{d\times d}, \ 
\mathrm{Cov}(\epsilon_{y,i},\epsilon_{y,j})=0, \quad \forall\, i \neq j,$}
\end{equation}
\begin{equation}
    \scalebox{0.95}{$\mathrm{Cov}(\epsilon_{x,i},\epsilon_{y,j}) = \mathbf{0}_{d\times 1}, \quad \forall\, i,j.$}
\label{eq:bias_noise_assumptions_cross}
\end{equation}

\subsection{Taylor Series Approximation with Bias Term}
\label{sec:bias_gp_taylor}
A first-order Taylor approximation about the recorded location \(\hat{\mathbf{x}}_i\) gives
\begin{equation}
h\big(\hat{\mathbf{x}}_i+\hat{\delta}_i+\epsilon_{x,i}\big)
\approx
h(\hat{\mathbf{x}}_i)
+
\nabla h(\hat{\mathbf{x}}_i)^{\top}\hat{\delta}_i
+
\nabla h(\hat{\mathbf{x}}_i)^{\top}\epsilon_{x,i},
\label{eq:bias_taylor_expansion}
\end{equation}
where \(\nabla h(\hat{\mathbf{x}}_i)\in\mathbb{R}^{d}\) denotes the gradient of \(h\) at \(\hat{\mathbf{x}}_i\).

\begin{assumption}
For each \B{ \(i=1,\dots,N\)}, the norm \(\|\hat{\delta}_i+\epsilon_{x,i}\|\) is small relative to the length-scale \(\ell\), so a first-order expansion about \(\hat{\mathbf{x}}_i\) is locally accurate.
\end{assumption}
Substituting \eqref{eq:bias_taylor_expansion} into \eqref{eq:bias_observation_compact} gives
\begin{equation}
y_i
\approx
h(\hat{\mathbf{x}}_i)
+
\nabla h(\hat{\mathbf{x}}_i)^{\top}\hat{\delta}_i
+
\nabla h(\hat{\mathbf{x}}_i)^{\top}\epsilon_{x,i}
+
\epsilon_{y,i}.
\label{eq:bias_linearized_model}
\end{equation}
Equation \eqref{eq:bias_linearized_model} separates the deterministic correction \(\nabla h(\hat{\mathbf{x}}_i)^{\top}\hat{\delta}_i\) from the zero-mean noise term \(\nabla h(\hat{\mathbf{x}}_i)^{\top}\epsilon_{x,i}\), which contribute to the mean and variance corrections, respectively.

\subsection{Approximate Likelihood Model}
\label{sec:bias_gp_moments}
\B{Following the NIGP approach \cite{SCC.McHutchon.Rasmussen2011} and 
replacing \(\nabla h(\hat{\mathbf{x}}_i)\) in \eqref{eq:bias_linearized_model} with \(\nabla\mu^{c}(\hat{\mathbf{x}}_i)\) gives
\begin{equation}
y_i
\approx
h(\hat{\mathbf{x}}_i)
+\nabla\mu^{c}(\hat{\mathbf{x}}_i)^{\top}\hat{\delta}_i
+\nabla\mu^{c}(\hat{\mathbf{x}}_i)^{\top}\epsilon_{x,i}
+\epsilon_{y,i}.
\label{eq:bias_plugin_model}
\end{equation}}Taking the expectation of \B{\eqref{eq:bias_plugin_model}} with respect to \(\epsilon_{x,i}\) and \(\epsilon_{y,i}\), conditioned on \(h, \hat{\mathbf{x}}_i\), and \(\hat{\delta}_i\), yields the approximate conditional mean
\begin{equation}
\mathbb{E}[y_i \mid h,\hat{\mathbf{x}}_i,\hat{\delta}_i]
\approx
h(\hat{\mathbf{x}}_i)
+
\hat{\delta}_i^{\top}\nabla \B{\mu^{c}}(\hat{\mathbf{x}}_i).
\label{eq:bias_mean_final}
\end{equation}
\B{Likewise, using \eqref{eq:bias_noise_assumptions_cross}, the approximate conditional variance is}
\begin{equation}
\mathrm{Var}[y_i \mid h,\hat{\mathbf{x}}_i,\hat{\delta}_i]
\approx
\nabla \B{\mu^{c}}(\hat{\mathbf{x}}_i)^{\top}\Sigma_{x,i}\nabla \B{\mu^{c}}(\hat{\mathbf{x}}_i)
+
\sigma_y^2.
\label{eq:bias_variance_final}
\end{equation}
Here, \(\mu^c(\cdot)\) and \((\sigma^c(\cdot))^2\) denote the \B{recorded GP} posterior mean and variance defined in \eqref{eq:gp_mean_corrupted} and
\eqref{eq:gp_variance_corrupted}. As in the NIGP approximation \cite{SCC.McHutchon.Rasmussen2011}, the random gradient is replaced by \(\nabla \mu^c(\hat{\mathbf{x}}_i)\), and derivative-covariance terms are neglected, yielding the diagonal correction in \eqref{eq:bias_variance_final}. To make the gradient term explicit, define
\(\B{\boldsymbol{\alpha}^{c}}\coloneqq (\B{K^{c,y}})^{-1}\B{\mathbf{Y}} \in \mathbb{R}^{N}\). From \eqref{eq:gp_mean_corrupted}, the gradient of the recorded posterior mean is $\nabla \B{\mu^{c}}(\mathbf{x})
=
\nabla\!\left(\B{\mathbf{k}^{c}}(\mathbf{x})^{\top}\B{\boldsymbol{\alpha}^{c}}\right)$. For the squared exponential kernel, $\nabla \B{\mu^{c}}$ can be written as
$\nabla \B{\mu^{c}}(\mathbf{x})
=
\frac{1}{\ell^{2}}
\sum_{i=1}^{\B{N}}\alpha_{i}^{c}\,
k(\hat{\mathbf{x}}_{i},\mathbf{x})(\hat{\mathbf{x}}_{i}-\mathbf{x})$, where \(\alpha_{i}^{c}\) denotes the \(i\)-th entry of \(\B{\boldsymbol{\alpha}^{c}}\). In \eqref{eq:bias_mean_final}--\eqref{eq:bias_variance_final}, the gradient is evaluated at \(\mathbf{x}=\hat{\mathbf{x}}_i\).

\subsection{Gaussian Approximation and Bias-Aware Prediction}
\label{sec:bias_gp_gaussian_approx}
\B{The first-order model contains the product of a random GP derivative and the localization error and is therefore not Gaussian. After replacing \(\nabla h(\hat{\mathbf{x}}_i)\) by the fixed recorded GP posterior-mean gradient \(\nabla\mu^c(\hat{\mathbf{x}}_i)\), the remaining stochastic term is Gaussian. Thus, for \(i=1,\dots,N\), the approximate conditional likelihood is Gaussian, given by 
\begin{multline} q\left( y_i \mid h(\hat{\mathbf{x}}_i), \hat{\mathbf{x}}_i, \hat{\delta}_i \right) \approx \mathcal{N}\Big( h(\hat{\mathbf{x}}_i) + \hat{\delta}_i^{\top}\nabla\mu^{c}(\hat{\mathbf{x}}_i), \\ \nabla\mu^{c}(\hat{\mathbf{x}}_i)^{\top} \Sigma_{x,i} \nabla\mu^{c}(\hat{\mathbf{x}}_i) + \sigma_y^2 \Big). \label{eq:bias_gp_gaussian} \end{multline} \B{Define}}
\begin{align}
\B{\tilde{\mu}_{\delta}}(\hat{\mathbf{x}}_i)
&\coloneqq \hat{\delta}_i^{\top}\nabla \B{\mu^{c}}(\hat{\mathbf{x}}_i), \text{\ and}\label{eq:bias_pointwise_mean_correction}\\
\B{\tilde{\Sigma}}(\hat{\mathbf{x}}_i)
&\coloneqq \nabla \B{\mu^{c}}(\hat{\mathbf{x}}_i)^{\top}\Sigma_{x,i}\nabla \B{\mu^{c}}(\hat{\mathbf{x}}_i).
\label{eq:bias_pointwise_corrections}
\end{align}

\B{Since the input-noise terms \(\epsilon_{x,i}\) are assumed uncorrelated
across measurements, collecting these corrections over the complete dataset yields}
$\B{\tilde{\Sigma}}(\B{\hat{\mathbf{X}}})
\coloneqq
\mathrm{diag}\Big(
\B{\tilde{\Sigma}}(\hat{\mathbf{x}}_{1}),\dots,\B{\tilde{\Sigma}}(\hat{\mathbf{x}}_{\B{N}})
\Big)\in\mathbb{R}^{\B{N \times N}}$, and $\B{\tilde{\mu}_{\delta}}(\B{\hat{\mathbf{X}}})
\coloneqq
\bigl[\B{\tilde{\mu}_{\delta}}(\hat{\mathbf{x}}_{i})\bigr]_{i=1}^{N}\in\mathbb{R}^{N}$. Let \({\mathbf{x}}_\ast\in\mathcal{X}\) denote a test location and define
\B{$\mathbf{k}_{\ast}^{c}
\coloneqq
\mathbf{k}^{c}(\mathbf{x}_\ast)$ and $k_{\ast\ast} \coloneqq k(\mathbf{x}_\ast,\mathbf{x}_\ast)$}.

The corrections \(\B{\tilde{\mu}_{\delta}}(\B{\hat{\mathbf{X}}})\) and \(\B{\tilde{\Sigma}}(\B{\hat{\mathbf{X}}})\) are computed from the recorded GP posterior \(\B{\mu^{c}}(\cdot)\) and then treated as fixed data-dependent corrections. The resulting joint GP model is
\begin{equation}
\scalebox{0.85}{$
q\!\left(
\begin{bmatrix}
\B{\mathbf{Y}} \\
h_\ast
\end{bmatrix}
\,\middle|\,
\B{\hat{\mathbf{X}}},
\mathbf{x}_\ast,
\B{\tilde{\boldsymbol{\mu}}_{\delta}},
\B{\tilde{\Sigma}}
\right)
\sim
\mathcal{N}\left(
\begin{bmatrix}
\B{\tilde{\mu}_{\delta}}(\B{\hat{\mathbf{X}}}) \\
0
\end{bmatrix},
\begin{bmatrix}
\B{K^{c,y}} + \B{\tilde{\Sigma}}(\B{\hat{\mathbf{X}}}) & \B{\mathbf{k}_{\ast}^{c}} \\
(\B{\mathbf{k}_{\ast}^{c}})^\top & k_{\ast\ast}
\end{bmatrix}
\right)
$}
\label{eq:bias_joint_model}
\end{equation}
where \(h_\ast\coloneqq h(\mathbf{x}_\ast)\). No correction is applied at the test point because \eqref{eq:bias_pointwise_corrections} is defined only for training inputs, and the predictor is formed for the latent field at the exact query location \({\mathbf{x}}_\ast\). The cross-covariance block is evaluated at the recorded inputs; derivative-kernel corrections due to input bias are neglected, as in NIGP-style approximations, to preserve the recorded GP conditioning form. Conditioning \eqref{eq:bias_joint_model} yields $q\left(
h_\ast \mid \B{\hat{\mathbf{X}}},\B{\mathbf{Y}},{\mathbf{x}}_\ast
\right)
=
\mathcal{N}\left(
\B{\mu^{\mathrm{corr}}_{\ast}},
\B{(\sigma^{\mathrm{corr}}_{\ast})^2}
\right)$, with
\begin{multline}
\B{\mu^{\mathrm{corr}}_{\ast}}
=
(\B{\mathbf{k}_{\ast}^{c}})^\top
\big(\B{K^{c,y}}+\B{\tilde{\Sigma}}(\B{\hat{\mathbf{X}}})\big)^{-1} 
\big(
\B{\mathbf{Y}}-\B{\tilde{\mu}_{\delta}}(\B{\hat{\mathbf{X}}})
\big),
\label{eq:bias_pred_mean}
\end{multline}
and
\begin{multline}
\B{(\sigma^{\mathrm{corr}}_{\ast})^2}
=
k_{\ast\ast}
-
(\B{\mathbf{k}_{\ast}^{c}})^\top
\big(\B{K^{c,y}}+\B{\tilde{\Sigma}}(\B{\hat{\mathbf{X}}})\big)^{-1}
\B{\mathbf{k}_{\ast}^{c}}.
\label{eq:bias_pred_var}
\end{multline}
Equations~\eqref{eq:bias_pred_mean} and \eqref{eq:bias_pred_var} define the bias-aware corrected GP predictor. In particular, the correction is computed from the \B{recorded GP} posterior and then used to form a locally corrected predictor.

\subsection{Bias Correction Strategies and Computational Cost}
\label{sec:retrospective_cost}
\B{Computation of the corrected GP in ~\eqref{eq:bias_pred_mean} and \eqref{eq:bias_pred_var} (referred to hereafter as \emph{full Ba-GP}) given the recorded GP (i.e., $K^{c,y}$ and $\boldsymbol{\alpha}^{c}$) requires inversion of an $N\times N$ matrix, which incurs $\mathcal{O}(N^3)$ operations \cite{SCC.Golub.Loan1996}. Since retraining the GP from scratch with corrected inputs (referred to hereafter as \emph{retrained GP}) also 
requires \(\mathcal{O}(N^3)\) operations, full Ba-GP does not provide a computational advantage over retraining. In the following, we use the Ba-GP framework to provide three alternative strategies that require \(\mathcal{O}(N^2)\) or less operations.
\subsubsection{Mean-only Ba-GP}
A lower-cost approximation of full Ba-GP retains only the deterministic mean correction while reusing the covariance and the kernel Gram matrix of the recorded GP, i.e.,
\begin{align}
\mu_{\ast}^{\mathrm{MO}}
&=
(\mathbf{k}_{\ast}^{c})^\top
(K^{c,y})^{-1}
\bigl(
\mathbf{Y}
-\tilde{\boldsymbol{\mu}}_{\delta}(\hat{\mathbf X})
\bigr),
\\
(\sigma_{\ast}^{\mathrm{MO}})^2
&=
(\sigma^c(\mathbf{x}_{\ast}))^2 .
\label{eq:bagp_mean_only}
\end{align}
Reusing the covariance and the inverse of ($K^{c,y}$) reduces the computational cost to \(\mathcal{O}(N^2)\).}

\subsubsection{First-Order Ba-GP Neumann Approximation}
\label{sec:bagp_neumann_first}
\B{
Full Ba-GP requires inversion of
\(K^{c,y}+\tilde{\Sigma}(\hat{\mathbf X})\). To avoid this inversion, we expand the
corrected inverse about the already available recorded GP covariance matrix
\(K^{c,y}\). Retaining the first-order term of the Neumann expansion gives
\begin{equation}
\tilde K_{\mathrm{Neu},1}^{-1}
=
(K^{c,y})^{-1}
-
(K^{c,y})^{-1}
\tilde{\Sigma}(\hat{\mathbf X})
(K^{c,y})^{-1}.
\label{eq:neumann1_inverse}
\end{equation}
Note that the inverse is already available from the recorded GP and is reused in this computation. Using the approximation \eqref{eq:neumann1_inverse} instead of the matrix inverse in the full Ba-GP predictor gives
\begin{align}
\mu_{\ast}^{\mathrm{Neu},1}
&=
(\mathbf{k}_{\ast}^{c})^\top
\tilde K_{\mathrm{Neu},1}^{-1}
\bigl(
\mathbf{Y}
-
\tilde{\boldsymbol{\mu}}_{\delta}(\hat{\mathbf X})
\bigr),
\label{eq:neumann1_bagp_mean}\\
(\sigma_{\ast}^{\mathrm{Neu},1})^2
&=
k(\mathbf{x}_{\ast},\mathbf{x}_{\ast})
-
(\mathbf{k}_{\ast}^{c})^\top
\tilde K_{\mathrm{Neu},1}^{-1}
\mathbf{k}_{\ast}^{c}.
\label{eq:neumann1_bagp_var}
\end{align}
Since \((K^{c,y})^{-1}\) is reused, 
the first-order update has a complexity of
\(\mathcal{O}(N^2)\).

Since \(K^{c,y}\succ0\) and
\(\tilde{\Sigma}(\hat{\mathbf X})\succeq0\), the normalized perturbation is
symmetric positive semidefinite. As a result, the Neumann series converges under the corresponding spectral-radius condition (cf. \cite[Sec.~5.6]{SCC.Horn.Johnson1993}).
\begin{equation}
\eta
=
\lambda_{\max}\!\left(
(K^{c,y})^{-1/2}
\tilde{\Sigma}(\hat{\mathbf X})
(K^{c,y})^{-1/2}
\right) < 1,
\label{eq:neumann_condition}
\end{equation} }

\subsubsection{Second-Order Ba-GP Neumann Approximation}
\label{sec:bagp_neumann_second}
\B{Including the second-order term of the Neumann expansion provides a higher-order approximation of the Ba-GP inverse
\begin{align}
\tilde K_{\mathrm{Neu},2}^{-1}
={}&
(K^{c,y})^{-1}
-
(K^{c,y})^{-1}
\tilde{\Sigma}(\hat{\mathbf X})
(K^{c,y})^{-1}
\nonumber\\
&+
(K^{c,y})^{-1}
\tilde{\Sigma}(\hat{\mathbf X})
(K^{c,y})^{-1}
\tilde{\Sigma}(\hat{\mathbf X})
(K^{c,y})^{-1}.
\label{eq:neumann2_inverse}
\end{align}
The corresponding predictive mean and variance are
\begin{align}
\mu_{\ast}^{\mathrm{Neu},2}
&=
(\mathbf{k}_{\ast}^{c})^\top
\tilde K_{\mathrm{Neu},2}^{-1}
\bigl(
\mathbf{Y}
-
\tilde{\boldsymbol{\mu}}_{\delta}(\hat{\mathbf X})
\bigr),
\label{eq:neumann2_bagp_mean}\\
(\sigma_{\ast}^{\mathrm{Neu},2})^2
&=
k(\mathbf{x}_{\ast},\mathbf{x}_{\ast})
-
(\mathbf{k}_{\ast}^{c})^\top
\tilde K_{\mathrm{Neu},2}^{-1}
\mathbf{k}_{\ast}^{c}.
\label{eq:neumann2_bagp_var}
\end{align}
Relative to the first-order approximation, the second-order form requires
one additional application of \((K^{c,y})^{-1}\) and one additional multiplication by \(\tilde{\Sigma}(\hat{\mathbf X})\). Since the number of such operations is fixed, the computational complexity remains \(\mathcal{O}(N^2)\).

The convergence condition in \eqref{eq:neumann_condition} ensures convergence of the Neumann series underlying the second-order approximation \cite{SCC.Horn.Johnson1993}. Thus, the second-order approximation reduces the
Neumann-series truncation error while retaining the same asymptotic complexity. More generally, an \(m\)-th order Neumann approximation is obtained by
retaining terms through order \(m\) in the expansion. Under
\eqref{eq:neumann_condition}, increasing the truncation order reduces the Neumann-series truncation error \cite{SCC.Horn.Johnson1993}. Table~\ref{tab:retrospective_cost} summarizes the four Ba-GP variants considered in this work.
}
\begin{table}[t]
\centering
\caption{Computational complexity of the Ba-GP variants.}
\label{tab:retrospective_cost}
\begingroup
\setlength{\tabcolsep}{12pt}
\renewcommand{\arraystretch}{0.95}
\begin{tabular}{@{}lccc@{}}
\toprule
Method & New matrix inverse & Cost \\
\midrule
Full Ba-GP                    & Yes & \(\mathcal{O}(N^3)\) \\
Mean-only Ba-GP               & No  & \(\mathcal{O}(N^2)\) \\
Ba-GP Neumann (first order)   & No  & \(\mathcal{O}(N^2)\) \\
Ba-GP Neumann (second order)  & No  & \(\mathcal{O}(N^2)\) \\
\bottomrule
\end{tabular}
\endgroup
\end{table}

\begin{table}[t]
\centering
\caption{Simulation comparison across methods.}
\label{tab:sim_metrics_norm}
\begingroup
\setlength{\tabcolsep}{11pt}
\renewcommand{\arraystretch}{0.9}

\begin{filecontents*}{V2_Simulation_data/gp_summary_raw.csv}
Method,RMSE_mean,RMSE_std,MAE-mean,MAE_std,NLPD-mean,NLPD_std,Cov95-mean,Cov95_std
Reference GP,0.149895785373189,0.0361385889493079,0.123038935162678,0.0315688445715697,-0.451873152220136,0.218336270490775,0.9585,0.0572414093761537
Recorded GP,0.244298894275435,0.0426005236996428,0.205685882499846,0.0356259545597107,0.251373314001208,0.36615510601845,0.8095,0.0878680230441325
Retrained GP,0.150884616305877,0.0345671662683891,0.12490297753539,0.0307969771957382,-0.447790237609005,0.209758436032747,0.972,0.0541537771438648
Dallaire GP,0.245054444677635,0.0425003174431425,0.206085704974723,0.0355314795302604,0.242551673694981,0.358214301327354,0.815,0.0861149782191473
NIGP,0.242450555871159,0.040898865897891,0.204780915395935,0.0351017778785386,0.178452839326363,0.30600944816807,0.834,0.0873830044899133
Full Ba-GP,0.157956102968779,0.028197340086671,0.131864753271236,0.0230727405525152,-0.406419494210328,0.160676158856486,0.9695,0.0437065573786673
Mean-only Ba-GP,0.159829624710905,0.0290520116149553,0.133454104422651,0.0235558231199797,-0.400382852712153,0.174250053700222,0.9645,0.0462800400309157
Ba-GP Neumann (first order),0.157493659549197,0.0279569668793752,0.131546057536938,0.0229754321891186,-0.408724155603606,0.159377158740774,0.9705,0.0427323370713438
Ba-GP Neumann (second order),0.158134019474036,0.0283392131574261,0.131966527152778,0.0231241880656574,-0.405431314919354,0.161359529624964,0.969,0.0443550270099414
\end{filecontents*}
\pgfplotstabletypeset[
col sep=comma,
columns={[index]0,[index]3,[index]5,[index]7},
display columns/0/.style={
    column name={Method},
    string type,
    column type={@{}l}
},
display columns/1/.style={
    column name={MAE \(\downarrow\)},
    column type=c,
    fixed,
    fixed zerofill,
    precision=4
},
display columns/2/.style={
    column name={NLPD \(\downarrow\)},
    column type=c,
    fixed,
    fixed zerofill,
    precision=4
},
display columns/3/.style={
    column name={\(\mathrm{Cov}_{95}\)},
    column type={c@{}},
    fixed,
    fixed zerofill,
    precision=4
},
every head row/.style={
    before row=\toprule,
    after row=\midrule
},
every row no 4/.style={
    after row={
        Offline Taylor series & 0.1442 & 1.8860 & 0.9225 \\
    }
},
every last row/.style={
    after row=\bottomrule
}
]{V2_Simulation_data/gp_summary_raw.csv}

\endgroup
\end{table}

\section{Numerical Study}
\label{sec:simulation}
\subsection{Simulation results}
The Ba-GP framework is evaluated on a 1D regression problem with known latent field \(h(x)=\sin(x)\). We generate \B{\(N=20\)} training samples from uniformly spaced true locations in \([-4,4]\). \B{The input--location error follows the model in
\eqref{eq:pf_posterior_input_model}, with the bias magnitude varying from \(0\) to \(0.5\) across the training locations and input noise with
variance \(0.01\).} The test set contains \B{\(N_{\mathrm{test}}=100\)} uniformly spaced test points in \([-4,4]\). 
\B{We compare ten methods for the simulation. The reference GP is a GP trained on the exact locations \(x_i\), while the recorded GP is trained on the recorded locations \(\hat{x}_i\). The \emph{retrained GP} is trained on the corrected inputs \(\hat{x}_i+\hat{\delta}_i\), which still
contain the residual noise. The Dallaire
\emph{et al.}~\cite{SCC.Dallaire.Besse.ea2009} method and
\emph{NIGP}~\cite{SCC.McHutchon.Rasmussen2011} are both trained on the recorded locations \(\hat{x}_i\). These methods account for the
stochastic input uncertainty but do not account for the input bias. The offline Taylor series technique recently published in ~\cite{SCC.Qureshi.Ogri.ECCea2026} exploits offline-computed Jacobians and Hessians of the recorded GP posterior with respect to the training locations to retrospectively correct the posterior using gradient descent. The four Ba-GP variants are
\emph{Full Ba-GP}, which applies both the
\(\tilde{\mu}_{\delta}\) and \(\tilde{\Sigma}\) corrections;
\emph{Mean-only Ba-GP}, which retains only the mean correction; and the first- and second-order \emph{Ba-GP Neumann} approximations of full Ba-GP.}

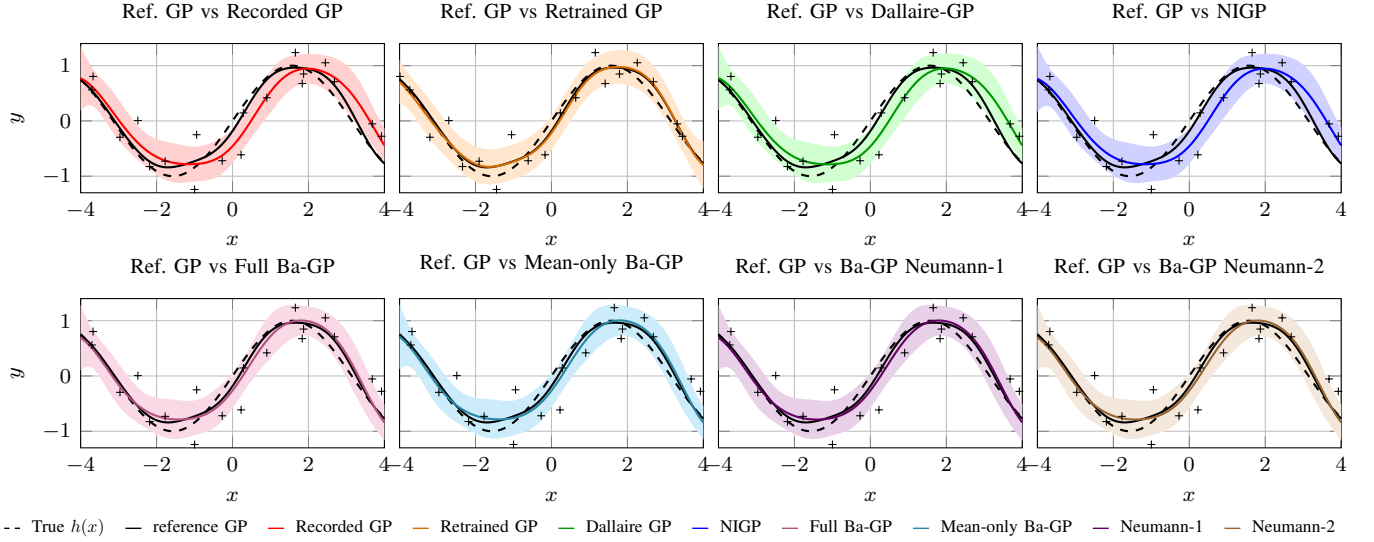
\begin{figure*}[t]
\centering
\begin{tikzpicture}

\begin{groupplot}[
    group style={
        group size=4 by 2,
        horizontal sep=0.20cm,
        vertical sep=1.40cm
    },
    width=0.315\textwidth,
    height=0.2\textwidth,
    grid=both,
    xlabel={$x$},
    ymin=-1.3,
    ymax=1.4,
    xmin=-4,
    xmax=4,
    tick label style={font=\footnotesize},
    label style={font=\footnotesize},
    title style={font=\footnotesize}
]

\nextgroupplot[
    title={Ref. GP vs Recorded GP},
    ylabel={$y$}
]

\addplot[black, dashed, thick]
table[x=X,y=y_true,col sep=comma]
{V2_Simulation_data/gp_predictions.csv};

\addplot[only marks, mark=+, mark size=1.5pt]
table[x=X_train,y=y_train,col sep=comma]
{V2_Simulation_data/gp_training.csv};

\addplot[black, thick]
table[x=X,y=mu_truegp,col sep=comma]
{V2_Simulation_data/gp_predictions.csv};

\addplot[name path=stdU, draw=none]
table[
    x=X,
    y expr=\thisrow{mu_standard}+2*\thisrow{sigma_standard},
    col sep=comma
]{V2_Simulation_data/gp_predictions.csv};

\addplot[name path=stdL, draw=none]
table[
    x=X,
    y expr=\thisrow{mu_standard}-2*\thisrow{sigma_standard},
    col sep=comma
]{V2_Simulation_data/gp_predictions.csv};

\addplot[red!18] fill between[of=stdU and stdL];

\addplot[red, thick]
table[x=X,y=mu_standard,col sep=comma]
{V2_Simulation_data/gp_predictions.csv};

\nextgroupplot[
    title={Ref. GP vs Retrained GP},
    ylabel={},
    yticklabels=\empty
]

\addplot[black, dashed, thick]
table[x=X,y=y_true,col sep=comma]
{V2_Simulation_data/gp_predictions.csv};

\addplot[only marks, mark=+, mark size=1.5pt]
table[x=X_corrected,y=y_train,col sep=comma]
{V2_Simulation_data/gp_training.csv};

\addplot[black, thick]
table[x=X,y=mu_truegp,col sep=comma]
{V2_Simulation_data/gp_predictions.csv};

\addplot[name path=retU, draw=none]
table[
    x=X,
    y expr=\thisrow{mu_retrained}+2*\thisrow{sigma_retrained},
    col sep=comma
]{V2_Simulation_data/gp_predictions.csv};

\addplot[name path=retL, draw=none]
table[
    x=X,
    y expr=\thisrow{mu_retrained}-2*\thisrow{sigma_retrained},
    col sep=comma
]{V2_Simulation_data/gp_predictions.csv};

\addplot[orange!20] fill between[of=retU and retL];

\addplot[orange!85!black, thick]
table[x=X,y=mu_retrained,col sep=comma]
{V2_Simulation_data/gp_predictions.csv};

\nextgroupplot[
    title={Ref. GP vs Dallaire-GP},
    ylabel={},
    yticklabels=\empty
]

\addplot[black, dashed, thick]
table[x=X,y=y_true,col sep=comma]
{V2_Simulation_data/gp_predictions.csv};

\addplot[only marks, mark=+, mark size=1.5pt]
table[x=X_train,y=y_train,col sep=comma]
{V2_Simulation_data/gp_training.csv};

\addplot[black, thick]
table[x=X,y=mu_truegp,col sep=comma]
{V2_Simulation_data/gp_predictions.csv};

\addplot[name path=dalU, draw=none]
table[
    x=X,
    y expr=\thisrow{mu_dallaire}+2*\thisrow{sigma_dallaire},
    col sep=comma
]{V2_Simulation_data/gp_predictions.csv};

\addplot[name path=dalL, draw=none]
table[
    x=X,
    y expr=\thisrow{mu_dallaire}-2*\thisrow{sigma_dallaire},
    col sep=comma
]{V2_Simulation_data/gp_predictions.csv};

\addplot[green!18] fill between[of=dalU and dalL];

\addplot[green!60!black, thick]
table[x=X,y=mu_dallaire,col sep=comma]
{V2_Simulation_data/gp_predictions.csv};

\nextgroupplot[
    title={Ref. GP vs NIGP},
    ylabel={},
    yticklabels=\empty
]

\addplot[black, dashed, thick]
table[x=X,y=y_true,col sep=comma]
{V2_Simulation_data/gp_predictions.csv};

\addplot[only marks, mark=+, mark size=1.5pt]
table[x=X_train,y=y_train,col sep=comma]
{V2_Simulation_data/gp_training.csv};

\addplot[black, thick]
table[x=X,y=mu_truegp,col sep=comma]
{V2_Simulation_data/gp_predictions.csv};

\addplot[name path=nigpU, draw=none]
table[
    x=X,
    y expr=\thisrow{mu_nigp}+2*\thisrow{sigma_nigp},
    col sep=comma
]{V2_Simulation_data/gp_predictions.csv};

\addplot[name path=nigpL, draw=none]
table[
    x=X,
    y expr=\thisrow{mu_nigp}-2*\thisrow{sigma_nigp},
    col sep=comma
]{V2_Simulation_data/gp_predictions.csv};

\addplot[blue!18] fill between[of=nigpU and nigpL];

\addplot[blue, thick]
table[x=X,y=mu_nigp,col sep=comma]
{V2_Simulation_data/gp_predictions.csv};

\nextgroupplot[
    title={Ref. GP vs Full Ba-GP},
    ylabel={$y$}
]

\addplot[black, dashed, thick]
table[x=X,y=y_true,col sep=comma]
{V2_Simulation_data/gp_predictions.csv};

\addplot[only marks, mark=+, mark size=1.5pt]
table[x=X_train,y=y_train,col sep=comma]
{V2_Simulation_data/gp_training.csv};

\addplot[black, thick]
table[x=X,y=mu_truegp,col sep=comma]
{V2_Simulation_data/gp_predictions.csv};

\addplot[name path=fullU, draw=none]
table[
    x=X,
    y expr=\thisrow{mu_full_bagp}+2*\thisrow{sigma_full_bagp},
    col sep=comma
]{V2_Simulation_data/gp_predictions.csv};

\addplot[name path=fullL, draw=none]
table[
    x=X,
    y expr=\thisrow{mu_full_bagp}-2*\thisrow{sigma_full_bagp},
    col sep=comma
]{V2_Simulation_data/gp_predictions.csv};

\addplot[magenta!18] fill between[of=fullU and fullL];

\addplot[magenta!70!black, thick]
table[x=X,y=mu_full_bagp,col sep=comma]
{V2_Simulation_data/gp_predictions.csv};

\nextgroupplot[
    title={Ref. GP vs Mean-only Ba-GP},
    ylabel={},
    yticklabels=\empty
]

\addplot[black, dashed, thick]
table[x=X,y=y_true,col sep=comma]
{V2_Simulation_data/gp_predictions.csv};

\addplot[only marks, mark=+, mark size=1.5pt]
table[x=X_train,y=y_train,col sep=comma]
{V2_Simulation_data/gp_training.csv};

\addplot[black, thick]
table[x=X,y=mu_truegp,col sep=comma]
{V2_Simulation_data/gp_predictions.csv};

\addplot[name path=meanU, draw=none]
table[
    x=X,
    y expr=\thisrow{mu_mean_bagp}+2*\thisrow{sigma_mean_bagp},
    col sep=comma
]{V2_Simulation_data/gp_predictions.csv};

\addplot[name path=meanL, draw=none]
table[
    x=X,
    y expr=\thisrow{mu_mean_bagp}-2*\thisrow{sigma_mean_bagp},
    col sep=comma
]{V2_Simulation_data/gp_predictions.csv};

\addplot[cyan!18] fill between[of=meanU and meanL];

\addplot[cyan!60!black, thick]
table[x=X,y=mu_mean_bagp,col sep=comma]
{V2_Simulation_data/gp_predictions.csv};

\nextgroupplot[
    title={Ref. GP vs Ba-GP Neumann-1},
    ylabel={},
    yticklabels=\empty
]

\addplot[black, dashed, thick]
table[x=X,y=y_true,col sep=comma]
{V2_Simulation_data/gp_predictions.csv};

\addplot[only marks, mark=+, mark size=1.5pt]
table[x=X_train,y=y_train,col sep=comma]
{V2_Simulation_data/gp_training.csv};

\addplot[black, thick]
table[x=X,y=mu_truegp,col sep=comma]
{V2_Simulation_data/gp_predictions.csv};

\addplot[name path=neu1U, draw=none]
table[
    x=X,
    y expr=\thisrow{mu_neumann1_bagp}
           +2*\thisrow{sigma_neumann1_bagp},
    col sep=comma
]{V2_Simulation_data/gp_predictions.csv};

\addplot[name path=neu1L, draw=none]
table[
    x=X,
    y expr=\thisrow{mu_neumann1_bagp}
           -2*\thisrow{sigma_neumann1_bagp},
    col sep=comma
]{V2_Simulation_data/gp_predictions.csv};

\addplot[violet!18] fill between[of=neu1U and neu1L];

\addplot[violet!80!black, thick]
table[x=X,y=mu_neumann1_bagp,col sep=comma]
{V2_Simulation_data/gp_predictions.csv};

\nextgroupplot[
    title={Ref. GP vs Ba-GP Neumann-2},
    ylabel={},
    yticklabels=\empty
]

\addplot[black, dashed, thick]
table[x=X,y=y_true,col sep=comma]
{V2_Simulation_data/gp_predictions.csv};

\addplot[only marks, mark=+, mark size=1.5pt]
table[x=X_train,y=y_train,col sep=comma]
{V2_Simulation_data/gp_training.csv};

\addplot[black, thick]
table[x=X,y=mu_truegp,col sep=comma]
{V2_Simulation_data/gp_predictions.csv};

\addplot[name path=neu2U, draw=none]
table[
    x=X,
    y expr=\thisrow{mu_neumann2_bagp}
           +2*\thisrow{sigma_neumann2_bagp},
    col sep=comma
]{V2_Simulation_data/gp_predictions.csv};

\addplot[name path=neu2L, draw=none]
table[
    x=X,
    y expr=\thisrow{mu_neumann2_bagp}
           -2*\thisrow{sigma_neumann2_bagp},
    col sep=comma
]{V2_Simulation_data/gp_predictions.csv};

\addplot[brown!18] fill between[of=neu2U and neu2L];

\addplot[brown!80!black, thick]
table[x=X,y=mu_neumann2_bagp,col sep=comma]
{V2_Simulation_data/gp_predictions.csv};

\end{groupplot}
\end{tikzpicture}

\vspace{-0.35em}

\begingroup
\setlength{\tabcolsep}{1pt}

\makebox[\textwidth][c]{%
\resizebox{\textwidth}{!}{%
\begin{tabular}{@{}c@{\;}l@{\hspace{0.8em}}
                    c@{\;}l@{\hspace{0.8em}}
                    c@{\;}l@{\hspace{0.8em}}
                    c@{\;}l@{\hspace{0.8em}}
                    c@{\;}l@{\hspace{0.8em}}
                    c@{\;}l@{\hspace{0.8em}}
                    c@{\;}l@{\hspace{0.8em}}
                    c@{\;}l@{\hspace{0.8em}}
                    c@{\;}l@{\hspace{0.8em}}
                    c@{\;}l@{}}

\raisebox{0.35ex}{
\tikz{\draw[black,dashed,thick] (0,0)--(0.32,0);}
}
& True \(h(x)\)

&
\raisebox{0.35ex}{
\tikz{\draw[black,thick] (0,0)--(0.32,0);}
}
& reference GP

&
\raisebox{0.35ex}{
\tikz{\draw[red,thick] (0,0)--(0.32,0);}
}
& Recorded GP

&
\raisebox{0.35ex}{
\tikz{\draw[orange!85!black,thick] (0,0)--(0.32,0);}
}
& Retrained GP

&
\raisebox{0.35ex}{
\tikz{\draw[green!60!black,thick] (0,0)--(0.32,0);}
}
& Dallaire GP

&
\raisebox{0.35ex}{
\tikz{\draw[blue,thick] (0,0)--(0.32,0);}
}
& NIGP

&
\raisebox{0.35ex}{
\tikz{\draw[magenta!70!black,thick] (0,0)--(0.32,0);}
}
& Full Ba-GP

&
\raisebox{0.35ex}{
\tikz{\draw[cyan!60!black,thick] (0,0)--(0.32,0);}
}
& Mean-only Ba-GP

&
\raisebox{0.35ex}{
\tikz{\draw[violet!80!black,thick] (0,0)--(0.32,0);}
}
& Neumann-1

&
\raisebox{0.35ex}{
\tikz{\draw[brown!80!black,thick] (0,0)--(0.32,0);}
}
& Neumann-2

\end{tabular}
}
}

\endgroup

\vspace{-0.45em}

\caption{Pairwise comparison of the eight GP regression methods
against the reference GP. The dashed black curve denotes the known latent
field \(h(x)\), while the solid black curve denotes the reference GP trained at
the exact input locations. Each colored curve denotes the corresponding
method and the shaded region its \(\pm2\sigma\) interval. The Retrained GP uses corrected training locations, whereas the recorded GP, Dallaire GP, NIGP, Full Ba-GP, Mean-only Ba-GP, and the first- and
second-order Ba-GP Neumann approximations operate on the recorded inputs.}
\label{fig:sim_group_predictions}
\end{figure*}

\begin{figure}[t]
\centering
\begin{tikzpicture}
\begin{axis}[
    width=\columnwidth,
    height=0.5\columnwidth,
    grid=major,
    xmode=log,
    ymode=log,
    xmin=50,
    xmax=335,
    ymin=0.55e-5,
    xlabel={\(N\)},
    ylabel={time (s)},
    tick label style={font=\footnotesize},
    label style={font=\small},
    legend style={
        font=\tiny,
        at={(0.00,1)},
        anchor=north west,
        draw=none,
        row sep=-4pt
    },
    scaled ticks=false
]
\def\shiftTwo{0.01}
\def\shiftThree{1.0}
\def\baseTwo{1e-7}
\def\baseThree{1e-10}
\addplot[
    red,
    thick,
    mark=o,
    mark size=1.7pt]
table[
    x=N,
    y=Upd_retrain,
    col sep=comma
]{V2_Simulation_data/runtime_scaling.csv};
\addlegendentry{Retrained GP}
\addplot[
    black,
    thick,
    mark=triangle*,
    mark size=1.7pt]
table[
    x=N,
    y=Upd_full_BaGP,
    col sep=comma
]{V2_Simulation_data/runtime_scaling.csv};
\addlegendentry{Full Ba-GP}
\addplot[
    blue,
    thick,
    mark=square*,
    mark size=1.5pt
]
table[
    x=N,
    y=Upd_mean_only,
    col sep=comma
]{V2_Simulation_data/runtime_scaling.csv};
\addlegendentry{Mean-only Ba-GP}
\addplot[
    magenta!70!black,
    thick,
    mark=diamond*,
    mark size=1.7pt
]
table[
    x=N,
    y=Upd_neumann,
    col sep=comma
]{V2_Simulation_data/runtime_scaling.csv};
\addlegendentry{Ba-GP Neumann-1}
\addplot[
    blue!60!black,
    dashed,
    thick,
    mark=none
]
table[
    x=N,
    y expr=\shiftTwo*\baseTwo*(\thisrow{N})^2,
    col sep=comma
]{V2_Simulation_data/runtime_scaling.csv};
\addlegendentry{$\mathcal{O}(N^2)$}
\addplot[
    red!70!black,
    dashdotted,
    thick,
    mark=none
]
table[
    x=N,
    y expr=\shiftThree*\baseThree*(\thisrow{N})^3,
    col sep=comma
]{V2_Simulation_data/runtime_scaling.csv};
\addlegendentry{$\mathcal{O}(N^3)$}
\end{axis}
\end{tikzpicture}
\vspace{-2mm}
\caption{Computational scaling of bias correction strategies with increasing \(N\).}
\label{fig:sim_time_comparison}
\end{figure}
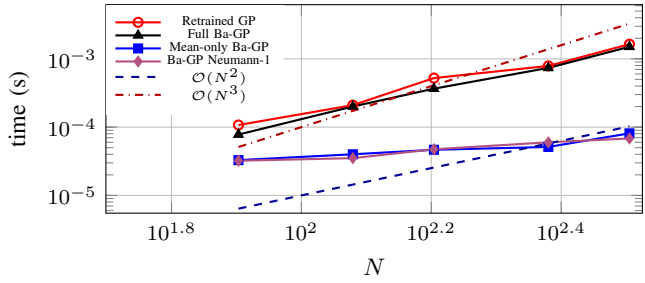

All methods use the hyperparameters as \((\ell,\sigma_f^2,\sigma_y^2)=(1,1,0.0625)\). For uncertain-input corrections, the input uncertainty is set to \(\sigma_x^2=0.01\). The performance of scalar-field mapping is evaluated using mean absolute error (MAE), negative log predictive density (NLPD), and empirical \(95\%\) coverage. Let \(h(x_j^\ast)\) be the true target and let \(\mu(x_j^\ast)\) and \((\sigma(x_j^\ast))^2\) be the predictive mean and predictive variance. The metrics are \(\mathrm{MAE}=\frac{1}{N_{\mathrm{test}}}\sum_j |h(x_j^\ast)-\mu(x_j^\ast)|\), \(\mathrm{NLPD}=\frac{1}{2N_{\mathrm{test}}}\sum_j\left[\log(2\pi(\sigma(x_j^\ast))^2)+\frac{(h(x_j^\ast)-\mu(x_j^\ast))^2}{(\sigma(x_j^\ast))^2}\right]\), and \(\mathrm{Cov}_{95}=\frac{1}{N_{\mathrm{test}}}\sum_j\mathbf{1}\!\left(|h(x_j^\ast)-\mu(x_j^\ast)|\le 1.96\,\sigma(x_j^\ast)\right)\). Lower MAE and NLPD are preferred, while \(\mathrm{Cov}_{95}\) is best interpreted relative to the nominal target \(0.95\). 
\B{Fig.~\ref{fig:sim_group_predictions} shows posterior means and \(\pm2\sigma\) bands, while Table~\ref{tab:sim_metrics_norm} reports metrics averaged over \B{20} seeds. The results show that correcting input bias brings the Retrained GP and Ba-GP variants substantially closer to the Reference GP than the Recorded GP, Dallaire GP, and NIGP, while the Neumann approximations closely match Full Ba-GP. Fig.~\ref{fig:sim_time_comparison} evaluates the computation time as the training dataset size increases. Its purpose is to demonstrate the computational distinction between Retrained GP and Full Ba-GP as given in Table~\ref{tab:retrospective_cost}.}

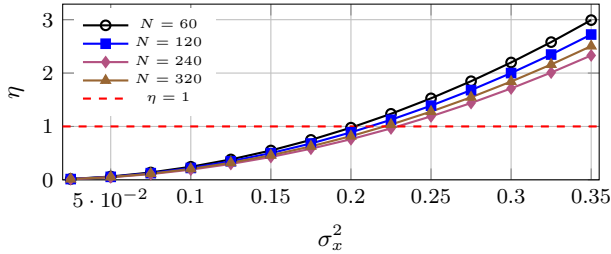
\begin{figure}[t]
\centering
\begin{tikzpicture}

\begin{axis}[
    width=\columnwidth,
    height=0.45\columnwidth,
    grid=major,
    xmin=0.02,
    xmax=0.355,
    ymin=0,
    xlabel={\(\sigma_x^2\)},
    ylabel={ \(\eta\)},
    tick label style={font=\footnotesize},
    label style={font=\small},
    legend style={
        font=\tiny,
        at={(0.02,0.98)},
        anchor=north west,
        draw=none,
        row sep=-2pt
    }
]
\addplot[
    black,
    thick,
    mark=o,
    mark size=1.8pt
]
table[
    x=sigma_resid,
    y=eta_N60,
    col sep=comma
]{V2_Simulation_data/neumann_analysis.csv};
\addlegendentry{\(N=60\)}

\addplot[
    blue,
    thick,
    mark=square*,
    mark size=1.6pt
]
table[
    x=sigma_resid,
    y=eta_N120,
    col sep=comma
]{V2_Simulation_data/neumann_analysis.csv};
\addlegendentry{\(N=120\)}

\addplot[
    magenta!70!black,
    thick,
    mark=diamond*,
    mark size=1.8pt
]
table[
    x=sigma_resid,
    y=eta_N240,
    col sep=comma
]{V2_Simulation_data/neumann_analysis.csv};
\addlegendentry{\(N=240\)}

\addplot[
    brown!80!black,
    thick,
    mark=triangle*,
    mark size=1.8pt
]
table[
    x=sigma_resid,
    y=eta_N320,
    col sep=comma
]{V2_Simulation_data/neumann_analysis.csv};
\addlegendentry{\(N=320\)}

\addplot[
    red,
    dashed,
    thick,
    mark=none
]
coordinates {
    (0.02,1)
    (0.355,1)
};
\addlegendentry{\(\eta=1\)}
\end{axis}
\end{tikzpicture}

\vspace{-2mm}
\caption{Neumann convergence metric
\(\eta\) as a function of input noise variance \(\sigma_x^2\) for different training-set sizes, while the dashed line denotes the Neumann-series
convergence boundary \(\eta=1\).}
\label{fig:neumann_convergence}
\end{figure}

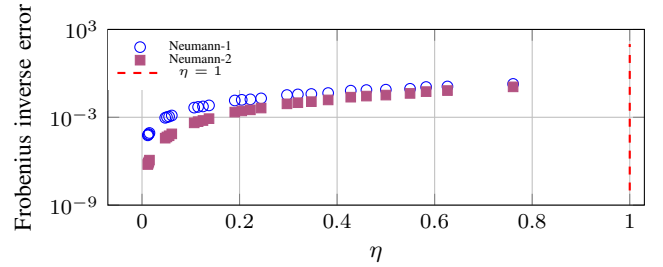
\begin{figure}[t]
\centering
\begin{tikzpicture}
\begin{axis}[
    width=\columnwidth,
    height=0.45\columnwidth,
    grid=major,
    ymode=log,
    xmin=-0.07,
    xmax=1.03,
    xlabel={ \(\eta\)},
    ylabel={Frobenius inverse error},
    tick label style={font=\footnotesize},
    label style={font=\small},
    legend style={
        font=\tiny,
        at={(0.0,1)},
        anchor=north west,
        draw=none,
        row sep=-4pt
    }
]
\addplot[
    blue,
    only marks,
    mark=o,
    mark size=2.0pt
]
table[
    x=eta_mean,
    y=E1_mean,
    col sep=comma
]{V2_Simulation_data/neumann_analysis_convergent.csv};
\addlegendentry{Neumann-1}

\addplot[
    magenta!70!black,
    only marks,
    mark=square*,
    mark size=1.8pt
]
table[
    x=eta_mean,
    y=E2_mean,
    col sep=comma
]{V2_Simulation_data/neumann_analysis_convergent.csv};
\addlegendentry{Neumann-2}

\addplot[
    red,
    dashed,
    thick,
    mark=none
]
coordinates {
    (1,1e-8)
    (1,1e2)
};
\addlegendentry{\(\eta=1\)}
\end{axis}
\end{tikzpicture}
\vspace{-2mm}
\caption{Relative Frobenius-norm error of the first- and second-order Neumann inverse approximations with respect to the exact Full Ba-GP inverse, plotted against the convergence metric \(\eta\).}
\label{fig:neumann_truncation_error}
\end{figure}

\B{In particular, the Retrained GP and Full Ba-GP exhibit the expected $\mathcal{O}(N^3)$ complexity, whereas the Mean-only Ba-GP and Neumann-1 approximations achieve $\mathcal{O}(N^2)$ complexity. Fig.~\ref{fig:neumann_convergence} shows the variation of $\eta$ with input noise variance $\sigma_x^2$ and $N$. The boundary $\eta = 1$ delineates the region where the Neumann-series convergence condition, $\eta < 1$, is satisfied. Fig.~\ref{fig:neumann_truncation_error} shows the relative Frobenius-norm inverse error of the Neumann-1 and Neumann-2 approximations as a function of the convergence metric $\eta$. The results quantify the truncation error introduced by finite-order Neumann-series approximations and demonstrate that the error increases as $\eta$ approaches the convergence boundary at $\eta=1$.}

\subsection{Experimental Results}
\label{subsec:Experimental}
\begin{figure}[h]
\centering
\begin{tikzpicture}
\node[inner sep=0] (lab) {\includegraphics[width=0.50\textwidth,height=0.1\textheight,]{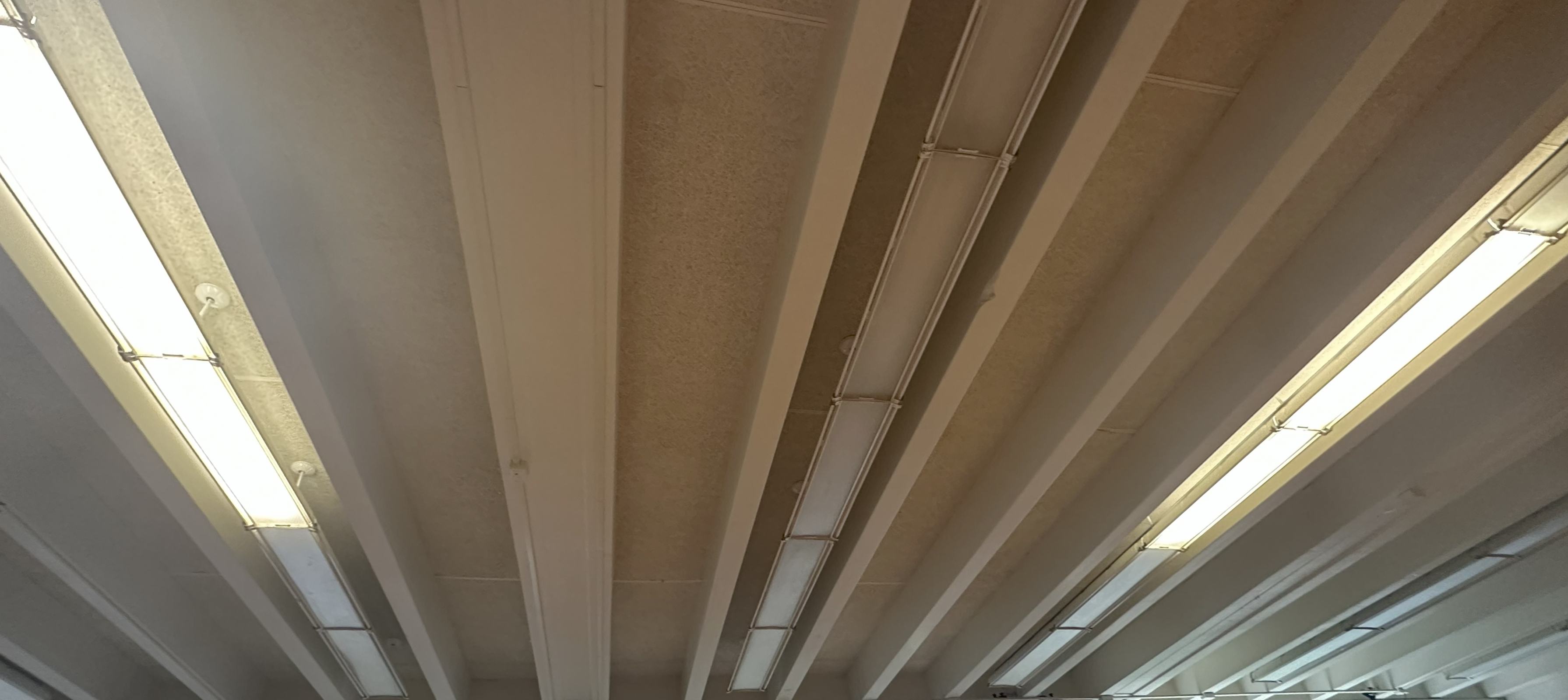}};
\node[
    draw=black,
    fill=white,
    inner sep=1pt,
    anchor=center
] at (lab.center)
{\includegraphics[width=0.11\textwidth]{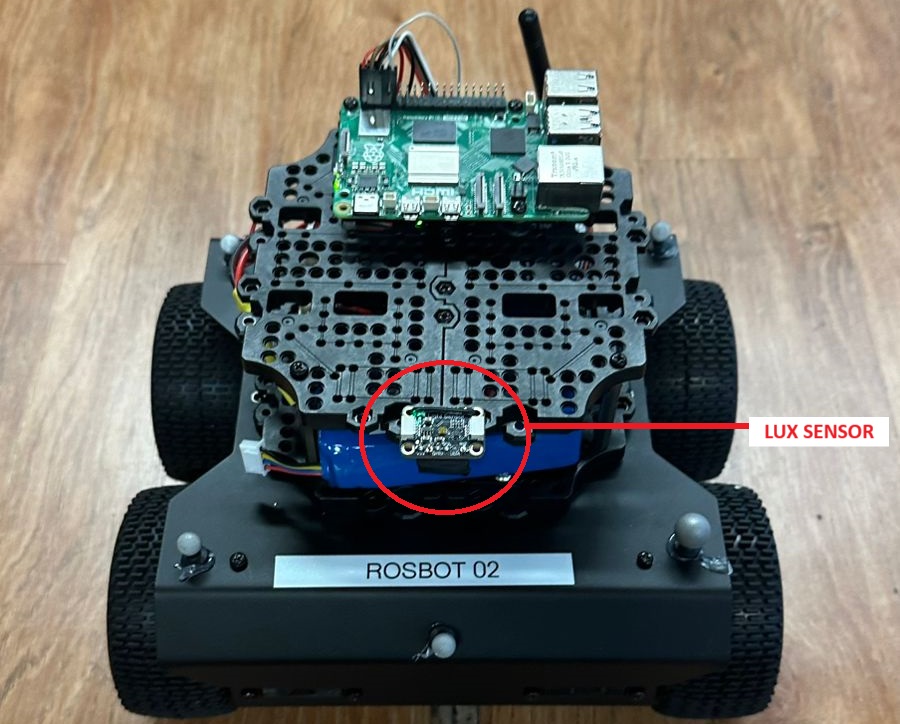}};
\end{tikzpicture}
\caption{Indoor light-mapping setup. The main image shows the ceiling-light arrangement, and the inset shows the Husarion ROSbot~2R Pro equipped with the Adafruit TSL2591 lux sensor used for data collection.}
\label{fig:lab_light}
\end{figure}

The Ba-GP framework is validated in an indoor light-intensity mapping experiment. The scalar field \(h\) represents ambient light intensity from the ceiling-light arrangement in Fig.~\ref{fig:lab_light}. Data are collected with a Husarion ROSbot~2R Pro equipped with an Adafruit TSL2591 lux sensor. \B{The recorded inputs are generated by the built-in localization filter by fusing wheel encoders with an inertial measurement unit. The corrected inputs are generated by fusing the recorded inputs with intermittent high-fidelity position updates (generated by a motion-capture system) using the following Kalman-filter/Rauch-Tung-Striebel (KF-RTS) procedure.}

\B{To obtain the bias estimates \(\hat{\delta}_i\) required by Ba-GP, the recorded robot trajectory is retrospectively refined using KF-RTS fixed-interval smoothing \cite{SCC.Rauch.Tung.ea1965}. Let \(\hat{p}(t_i)\in\mathbb{R}^{d}\) denote the position estimate used to associate the \(i\)-th field measurement with the recorded input \(\hat{\mathbf{x}}_i\), and let \(\hat{p}^{s}(t_i)\in\mathbb{R}^{d}\) denote the corresponding smoothed position. The estimated input bias is then
\begin{equation}
\hat{\delta}_i
\coloneqq
\hat{p}^{s}(t_i)-\hat{p}(t_i),
\qquad i=1,\dots,N.
\label{eq:delta_definition}
\end{equation}
Accordingly, the bias-corrected input location is
\begin{equation}
\hat{\mathbf{x}}_i+\hat{\delta}_i
=
\hat{p}^{s}(t_i).
\label{eq:refined_input_definition}
\end{equation}
The resulting \(\hat{\delta}_i\) are used as the bias
estimates in the Ba-GP formulation.}

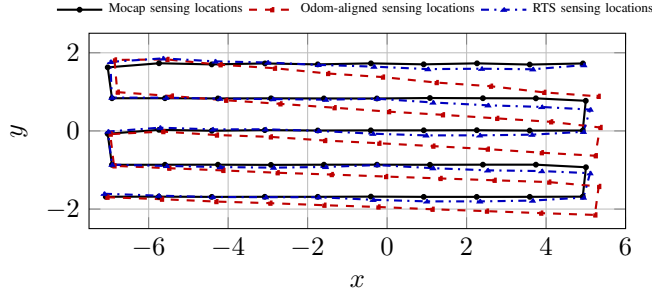
\begin{figure}
\centering
\begin{tikzpicture}
\begin{axis}[
    width=\linewidth,
    height=0.48\linewidth,
    grid=both,
    xlabel={$x$},
    ylabel={$y$},
    enlargelimits=false,
    xmin=-7.5,
    xmax=6,
    ymin=-2.5,
    ymax=2.5,
    legend style={
        font=\tiny,
        at={(0.5,1.03)},
        anchor=south,
        draw=none,
        fill=white,
        legend columns=3
    }
]

\addplot[
    black,
    thick,
    mark=*,
    mark size=0.7pt
]
table[
    x=mocap_x,
    y=mocap_y,
    col sep=comma
]
{V2_Experiment_data/experimental_training_data.csv};
\addlegendentry{Mocap sensing locations}

\addplot[
    red!75!black,
    thick,
    dashed,
    mark=square*,
    mark size=0.6pt
]
table[
    x=odom_aligned_x,
    y=odom_aligned_y,
    col sep=comma
]
{V2_Experiment_data/experimental_training_data.csv};
\addlegendentry{Odom-aligned sensing locations}

\addplot[
    blue!75!black,
    thick,
    dash dot,
    mark=triangle*,
    mark size=0.8pt
]
table[
    x=rts_x,
    y=rts_y,
    col sep=comma
]
{V2_Experiment_data/experimental_training_data.csv};
\addlegendentry{RTS sensing locations}

\end{axis}
\end{tikzpicture}

\caption{Robot trajectory during light mapping showing the motion-capture
reference sensing locations, aligned odometry sensing locations, and
RTS-smoothed sensing locations.}
\label{fig:exp_trajectory}
\end{figure}

\begin{figure}
\centering
\includegraphics[width=0.5\textwidth,height=0.1\textheight]{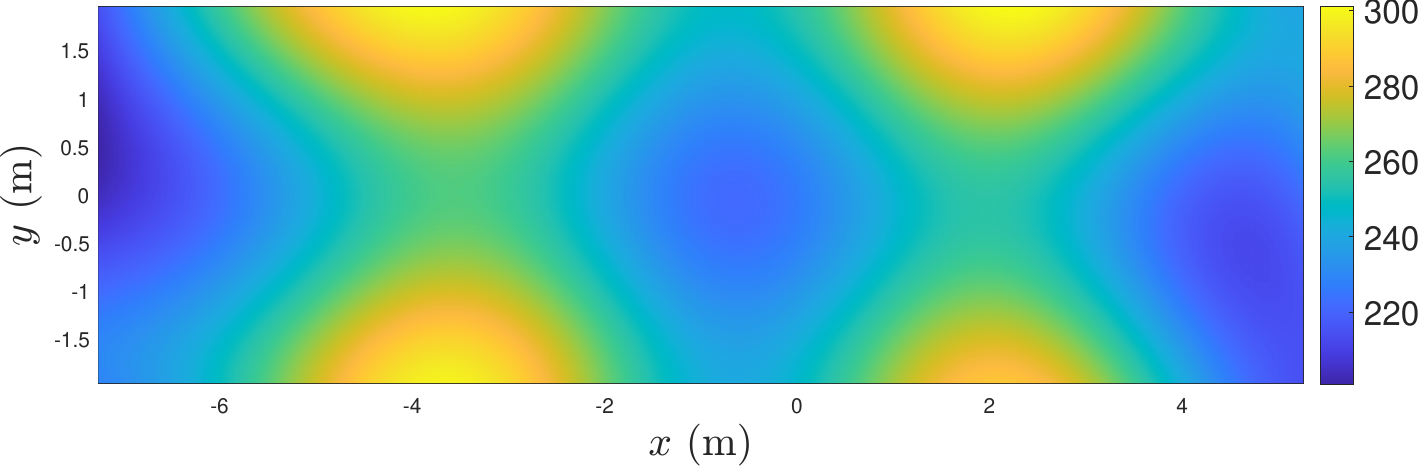}
\caption{Measured ground-level light-intensity field corresponding to the lighting layout in Fig.~\ref{fig:lab_light}.}
\label{fig:ground_lab_light}
\end{figure}

The KF-RTS procedure yields smoothed positions and bias estimates as shown in Fig.~\ref{fig:exp_trajectory}. These bias estimates are incorporated into the developed Ba-GP model to obtain bias-corrected predictions of the light-intensity field. \B{The GP prediction accuracy is evaluated against a reference light-intensity field constructed using inputs generated by a motion-capture system over repeated room-mapping trials as shown in Fig.~\ref{fig:ground_lab_light}.}

\B{The reference GP is trained using the inputs generated by the motion-capture system. The recorded GP is trained on the position estimates from the built-in localization filter. The retrained GP and Ba-GP variants are defined as in the simulation study. Fig.~\ref{fig:exp_heatmaps} compares the prediction errors of the recorded GP and Full Ba-GP relative to the reference light-intensity field. The recorded GP error map is overlaid with the odometry trajectory, whereas the Full Ba-GP error map is overlaid with the RTS-smoothed trajectory used to obtain the bias estimates.
}

Table~\ref{tab:exp_metrics} summarizes the quantitative experimental results. Full Ba-GP achieves lower MAE than the recorded GP, Dallaire GP, and NIGP baselines. The first- and second-order Neumann approximations closely match
Full Ba-GP across all reported metrics, with only negligible numerical differences in MAE, NLPD, and \(\mathrm{Cov}_{95}\). These results indicate that the truncated Neumann approximations retain the predictive performance of the
full Ba-GP formulation in this experiment. The experimental video is available at \href{https://youtu.be/CKP7FaPYd3Y}{https://youtu.be/CKP7FaPYd3Y}.

\begin{table}[t]
\caption{Experimental comparison across methods.}
\label{tab:exp_metrics}
\begingroup
\setlength{\tabcolsep}{11pt}
\renewcommand{\arraystretch}{0.9}
\begin{filecontents*}{V2_Experiment_data/experimental_gp_metrics.csv}
Method,MAE,NLPD,Cov95
Reference GP,4.88640261371165,3.30141889484603,0.979915278995034
Recorded GP,6.46088380855826,3.50524962607351,0.943616710487876
Retrained GP,4.71886217550021,3.27041133664897,0.984346090174311
Dallaire GP,6.46935168352989,3.50232878311693,0.948680494692765
NIGP,6.43198223643929,3.49121214540588,0.949508228649333
Full Ba-GP,4.78835316542008,3.30714101571868,0.982812347843023
Mean-only Ba-GP,4.7890196152891,3.30593462967654,0.982155029701042
Ba-GP Neumann (first order),4.78854900757527,3.30717753508807,0.982763657610283
\end{filecontents*}
\pgfplotstabletypeset[
col sep=comma,
columns={Method,MAE,NLPD,Cov95},
columns/Method/.style={
    column name={Method},
    string type,
    column type={@{}l}
},
columns/MAE/.style={
    column name={MAE \(\downarrow\)},
    column type=c,
    fixed,
    fixed zerofill,
    precision=4
},
columns/NLPD/.style={
    column name={NLPD \(\downarrow\)},
    column type=c,
    fixed,
    fixed zerofill,
    precision=4
},
columns/Cov95/.style={
    column name={\(\mathrm{Cov}_{95}\)},
    column type={c@{}},
    fixed,
    fixed zerofill,
    precision=4
},
every head row/.style={
    before row=\toprule,
    after row=\midrule
},
every last row/.style={
    after row=\bottomrule
}
]{V2_Experiment_data/experimental_gp_metrics.csv}

\endgroup
\end{table}
\arrayrulecolor{black}

\section{Discussion}
\label{sec:discussion}
A key implication of Ba-GP is that input bias and input noise should be treated separately. The derivation shows that input bias and input noise enter the GP correction differently: the former appears as a mean correction, whereas the latter contributes to the variance. This distinction explains why recorded GP degrades under localization drift and why variance-only noisy-input corrections such as NIGP are insufficient. The developed framework is modular, since any trajectory-refinement or localization method that provides corrected inputs and the associated posterior noise covariance could be incorporated, provided the independence assumption mentioned in Remark \ref{rem:independence} remains reasonable. For larger datasets, sparse GP approximations can be used to reduce computational cost while preserving the same bias-aware correction structure. \B{The results show that Ba-GP separates the correction of input bias from residual noise, while the developed approximation techniques provide computationally efficient alternatives to Full Ba-GP.}

\section{Conclusion}
\label{sec:conclusion}
This letter developed Ba-GP, a bias-aware Gaussian process framework for regression with \B{recorded} input locations containing \B{bias and noise}. By combining KF--RTS smoothing with a first-order GP correction, the method accounts for deterministic bias through a mean correction and for residual \B{noise} through an additional variance term. Results show improved predictive accuracy and uncertainty calibration over recorded GP and noisy-input baselines. \B{GP correction via Full Ba-GP and retraining using corrected inputs both have a  \(O(N^3)\) computational cost, while the Mean-only and Neumann approximations reuse the recorded GP covariance matrix and reduce the correction cost to \(O(N^2)\).}

\begin{center}
\begin{minipage}{0.98\columnwidth}
\centering

\includegraphics[
    width=\linewidth,
    height=0.22\textheight
]{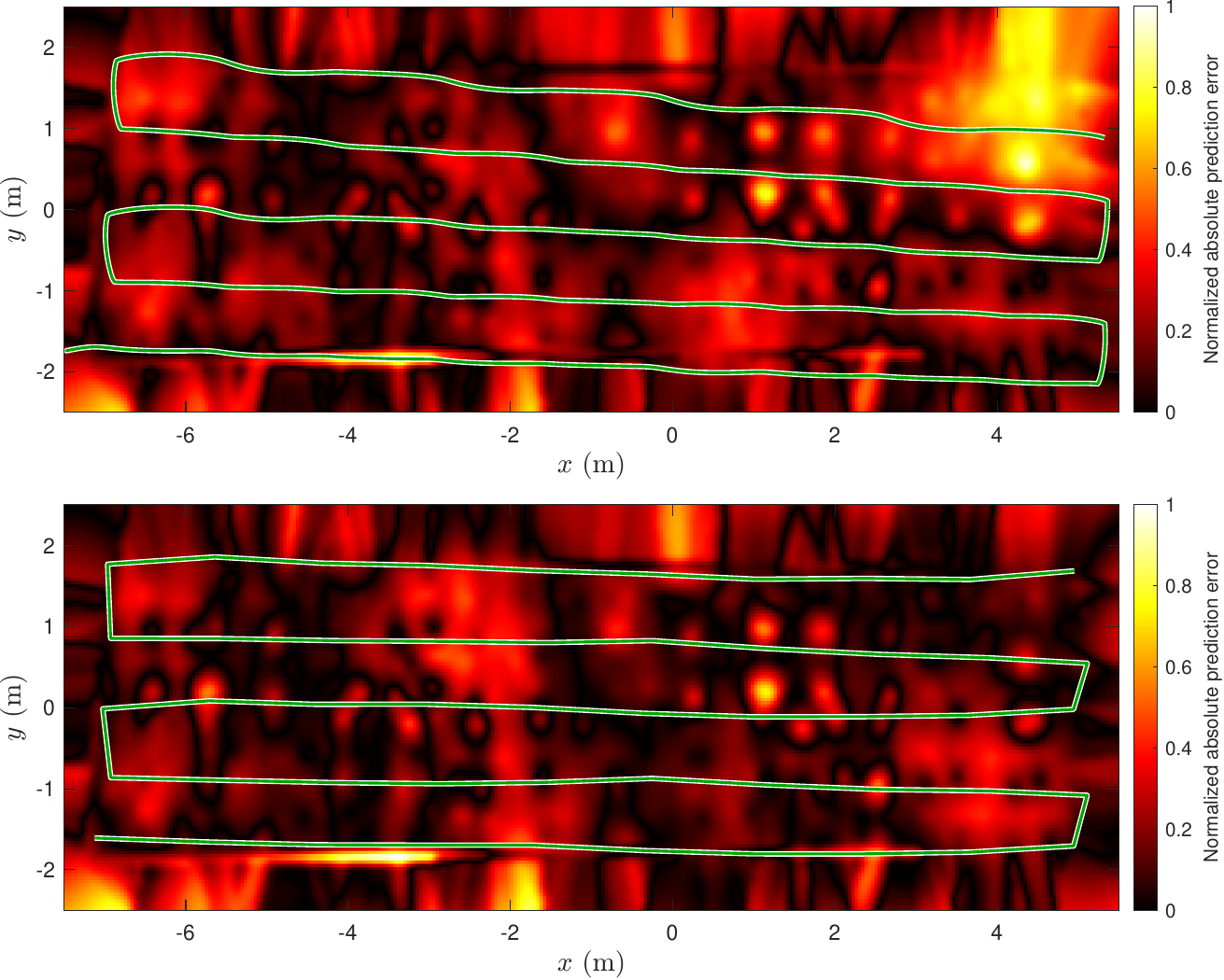}
\makeatletter
\def\@captype{figure}
\makeatother
\caption{Normalized absolute prediction-error heat maps relative to the
reference light-intensity field. The top panel shows the recorded GP
prediction error with the odometry trajectory overlaid, while the bottom
panel shows the Full Ba-GP prediction error with the RTS-smoothed
trajectory overlaid.}
\label{fig:exp_heatmaps}

\end{minipage}
\end{center}

\bibliographystyle{IEEEtran}
\bibliography{scc,sccmaster}

@article{Qureshi2025FaultTolerant,
  author    = {Qureshi, Muzaffar. and Maqsood, A. and Fayyaz ud Din, A.},
  title     = {Fault-Tolerant Control of a Quadcopter Using Reinforcement Learning},
  journal   = {SAE International Journal of Aerospace},
  volume    = {18},
  pages     = {75--89},
  year      = {2025},
  doi       = {10.4271/01-18-01-0006},
}

@misc{qureshi2025gaussianprocessbasedscalarfield,
      title={Gaussian Process-Based Scalar Field Estimation in {GPS-}Denied Environments}, 
      author={Qureshi, Muzaffar and Tochukwu Elijah Ogri and Humberto Ramos and Zachary I. Bell and Rushikesh Kamalapurkar},
      year={2025},
      eprint={2502.17584},
      archivePrefix={arXiv},
      primaryClass={eess.SY},
      howpublished = {{arXiv}:2502.17584}
}

@misc{TCST,
title={A {H}ough transform approach to safety-aware scalar field mapping using {G}aussian Processes}, 
author={Qureshi, Muzaffar and Ogri, Tochukwu E. and Wanjiku A. Makumi and Kyle Volle and Kamalapurkar, Rushikesh},
year = {2026},
eprint={arXiv:2604.20799},
archivePrefix={arXiv},
howpublished = {{arXiv}:2604.20799}
}

@Article{SCC.Qureshi.Ogri.ea2025,
  author       = {Qureshi, Muzaffar and Ogri, Tochukwu E. and Ramos, Humberto and Makumi, Wanjiku A. and Bell, Zachary I. and Kamalapurkar, Rushikesh},
  journal      = {IEEE Control Syst. Lett.},
  title        = {A switched adaptive control approach to reduce sensing needs in trajectory tracking problems},
  year         = {2025},
  month        = dec,
  pages        = {2777--2782},
  volume       = {9},
  doi          = {10.1109/LCSYS.2025.3641525},
  projects     = {afrl-fa8651-23-1-0006,afrl-fa8651-24-1-0019,onr-n00014-21-1-2481},
  url_code     = {https://codeocean.com/capsule/5400510/tree},
  url_preprint = {https://scc-lab.github.io/Preprints/},
}

@Book{SCC.Golub.Loan1996,
  title     = {Matrix computations},
  publisher = {The John Hopkins University Press},
  year      = {1996},
  author    = {Golub, Gene H. and van Loan, Charles F.},
  edition   = {Third},
}

@Book{SCC.Horn.Johnson1993,
  title     = {Matrix analysis},
  publisher = {Cambridge University Press.},
  year      = {1993},
  author    = {Horn, Roger A. and Johnson, Charles R.},
  address   = {Cambridge},
}

@Book{SCC.Rasmussen.Williams2006,
  author    = {Rasmussen, Carl Edward and Williams, Christopher K. I.},
  publisher = {MIT Press},
  title     = {{G}aussian processes for machine learning.},
  year      = {2006},
  address   = {Cambridge, MA},
  doi       = {10.7551/mitpress/3206.001.0001},
}

@InProceedings{SCC.Liu.Mao.ea2025,
  author    = {Liu, Shuo and Mao, Yihui and Belta, Calin A.},
  booktitle = {Proc. Amer. Control Conf.},
  title     = {Safety-Critical Planning and Control for Dynamic Obstacle Avoidance Using Control Barrier Functions},
  year      = {2025},
  pages     = {348--354},
  doi       = {10.23919/ACC63710.2025.11107805},
}

@Article{SCC.GhaffariJadidi.Miro.ea2017,
  author  = {Ghaffari Jadidi, Maani and Miro, Jaime Valls and Dissanayake, Gamini},
  journal = {IEEE Robot. Autom. Lett.},
  title   = {Warped {G}aussian Processes Occupancy Mapping With Uncertain Inputs},
  year    = {2017},
  number  = {2},
  pages   = {680--687},
  volume  = {2},
  doi     = {10.1109/LRA.2017.2651154},
}

@Article{SCC.Lilienthal.Duckett2004,
  author  = {Lilienthal, Achim and Duckett, Tom},
  journal = {Robot. Auton. Syst.},
  title   = {Building Gas Concentration Gridmaps with a Mobile Robot},
  year    = {2004},
  number  = {1},
  pages   = {3--16},
  volume  = {48},
  doi     = {https://doi.org/10.1016/j.robot.2004.05.002},
}

@InProceedings{SCC.Girard.Rasmussen.ea2002,
  author    = {Girard, Agathe and Rasmussen, Carl Edward and Candela, Joaquin Qui{\~n}onero and Murray-Smith, Roderick},
  booktitle = {Proc. Adv. Neural Inf. Process. Syst.},
  title     = {Gaussian Process Priors with Uncertain Inputs Application to Multiple-Step Ahead Time Series Forecasting},
  year      = {2002},
  volume    = {15},
}

@PhdThesis{SCC.Girard2004,
  author = {Girard, Agathe},
  school = {University of Glasgow},
  title  = {Approximate Methods for Propagation of Uncertainty with {G}aussian Process models},
  year   = {2004},
}

@InProceedings{SCC.Dallaire.Besse.ea2009,
  author    = {P. Dallaire and C. Besse and B. Chaib-draa},
  booktitle = {Proc. Int. Conf. Neural Inf. Process.},
  title     = {Learning {G}aussian Process Models from Uncertain Data},
  year      = {2009},
  pages     = {433--440},
  publisher = {Springer},
  doi       = {10.1007/978-3-642-10677-4_49},
}

@InProceedings{SCC.McHutchon.Rasmussen2011,
  author    = {McHutchon, Andrew and Rasmussen, Carl},
  booktitle = {Proc. Adv. Neural Inf. Proc. Syst.},
  title     = {Gaussian Process Training with Input Noise},
  year      = {2011},
  volume    = {24},
}

@InProceedings{SCC.Qureshi.Ogri.ECCea2026,
  author    = {Qureshi, Muzaffar and Ogri, Tochukwu Elijah and Volle, Kyle and Kamalapurkar, Rushikesh},
  booktitle = {Proc. Eur. Control Conf.},
  title     = {A {T}aylor Series Approach to Correct Localization Errors in Robotic Field Mapping using {G}aussian Processes},
  year      = {2026},
  pages     = {2045-2050},
}

@InProceedings{SCC.Julier.Uhlmann1997,
  author    = {Julier, S. J. and Uhlmann, J. K.},
  booktitle = {Proc. Am. Control Conf.},
  title     = {A Non-Divergent Estimation Algorithm in the Presence of Unknown Correlations},
  year      = {1997},
  pages     = {2369--2373},
  volume    = {4},
  doi       = {10.1109/ACC.1997.609105},
}

@Article{SCC.Rauch.Tung.ea1965,
  author  = {Rauch, H. E. and Tung, F. and Striebel, C. T.},
  journal = {AIAA J.},
  title   = {Maximum likelihood estimates of linear dynamic systems},
  year    = {1965},
  month   = aug,
  number  = {8},
  pages   = {1445--1450},
  volume  = {3},
  doi     = {10.2514/3.3166},
}

\end{document}